\documentclass{sig-alternate}

\usepackage{amsmath}
\usepackage{amssymb}
\usepackage{upquote}
\usepackage{url}
\usepackage{graphicx}

\newcommand\R{{\mathbb R}}
\newcommand\N{{\mathbb N}}
\DeclareMathOperator*\argmax{\operatorname{argmax}}

\usepackage{algorithm}
\usepackage{algorithmic}

\usepackage{booktabs}
\usepackage{multirow}
\usepackage{tabularx}
\usepackage[table]{xcolor}
\newcolumntype{C}[1]{>{\raggedleft\let\newline\\\arraybackslash\hspace{0pt}}m{#1}}

\usepackage{alltt}

\begin{document}
%
\conferenceinfo{XXX}{XXXX XXXXXXXXXXXXXXXX}

\title{Optimal Time-Series Motifs} 
%
%
%
%
%

\numberofauthors{3} 
%
\author{
%
%
\alignauthor
Josif Grabocka\\
       \affaddr{Information Systems and Machine Learning Lab }\\
       \affaddr{Hildesheim, Germany}\\
       \email{josif@ismll.de}
\alignauthor
Nicolas Schilling\\
       \affaddr{Information Systems and Machine Learning Lab }\\
       \affaddr{Hildesheim, Germany}\\
       \email{schilling@ismll.de}
\alignauthor Lars Schmidt-Thieme\\
       \affaddr{Information Systems and Machine Learning Lab }\\
       \affaddr{Hildesheim, Germany}\\
       \email{schmidt-thieme@ismll.de}
}

\maketitle
\begin{abstract}

Motifs are the most repetitive/frequent patterns of a time-series. The discovery of motifs is crucial for practitioners in order to understand and interpret the phenomena occurring in sequential data. Currently, motifs are searched among series sub-sequences, aiming at selecting the most frequently occurring ones. Search-based methods, which try out series sub-sequence as motif candidates, are currently believed to be the best methods in finding the most frequent patterns.

However, this paper proposes an entirely new perspective in finding motifs. We demonstrate that searching is non-optimal since the domain of motifs is restricted, and instead we propose a principled optimization approach able to find optimal motifs. We treat the occurrence frequency as a function and time-series motifs as its parameters, therefore we \textit{learn} the optimal motifs that maximize the frequency function. In contrast to searching, our method is able to discover the most repetitive patterns (hence optimal), even in cases where they do not explicitly occur as sub-sequences. Experiments on several real-life time-series datasets show that the motifs found by our method are highly more frequent than the ones found through searching, for exactly the same distance threshold. 

\end{abstract}




\section{Introduction}

Time-series are arguably the most widespread type of data which occur in virtually all the application domains of our modern lives, wherever measurements have associated time stamps (e.g.: physiological and medical, financial, meteorological, sound and video, monitoring system sensors, astronomy light intensities, and many more ). 

In many cases, the underlying patterns of those datasets are not known to the domain practitioners and a visual inspection is often infeasible given the complexity and size of the data. For this reason, finding the most repetitive patterns in time-series help the domain experts understand the underlying phenomena within diverse sources of data ~\cite{Buhler:2001:FMU:369133.369172,Syed:2010:MDP:1644873.1644875}. The most repetitive time-series patterns are called \textit{motifs} and their discovery has recently attracted considerable research~\cite{Patel:2002:MMM:844380.844710,10.1109/ICDM.2013.27,6729632,DBLP:conf/sdm/LiLO12}. In brief terms, optimal motifs are those which repeat the most (i.e. have the highest frequency) given a distance/similarity threshold value. The approach of the current state-of-the-art motif discovery methods is to \textit{\textbf{search}} the motifs from the segments (a.k.a sub-sequences) of time series~\cite{Patel:2002:MMM:844380.844710,Yankov:2007:DTS:1281192.1281282,DBLP:conf/sdm/LiLO12,MrMotif}. More concretely, series segments are considered to be motif candidates and the most frequent segments are sorted out.

In this paper we present an entirely new and orthogonally different perspective to the \textit{\textbf{search-based}} approach. First of all, we treat frequency as a function and motifs as its variable. Naturally our task becomes finding the values of motifs which maximize the value of the frequency function. In this perspective we formalize motif discovery as a principled optimization problem and devise an optimization technique to \textbf{\textit{learn}} the optimal motifs. The learning process uses the first order derivative of the frequency function, in order to find its maximum. In that way, our method can learn motifs which yield the maximum frequency (a.k.a the highest number of matches). The proposed \textbf{\textit{learning}} method is theoretically superior to the \textit{\textbf{search-based}} approach, because in the case of searching the motif candidates are limited to the domain of sub-sequences and cannot discover latent series patterns (Section~\ref{sec:motivation}) . 

As the empirical results (Section~\ref{sec:results}) over various real-life datasets will indicate, our optimal motifs have significantly more matches (higher frequency) than the ones found through searching, for exactly the same distance threshold.

\section{Related Work}
\label{sec:relatedWork}

The research on discovering time-series motifs has suffered from a terminological ambiguity. Initially, motifs were defined to be the most frequently occurring patterns in a time-series \cite{Patel:2002:MMM:844380.844710}. However, another stream of papers redefined the term "motif" as the closest pair among series segments \cite{mueen2009exact,Mueen:2010:ODM:1835804.1835941}. In this paper we mean "the most frequently occurring patterns" \cite{Patel:2002:MMM:844380.844710} when referring to motifs. The closest pair of series segments, on the other hand, will be referred to as "pair-motif" following the suggestion of \cite{yasser2014}.

\subsection{Pair-motif discovery}

The closest pair of series segments can be perceived as a sub-variation of the general motif discovery task. The brute-force search that computes the distance of every segment pair is computationally expensive, therefore efforts are devoted towards scaling the brute force up. A fast, yet exact, method that discovers pair-wise motifs has been introduced by~\cite{mueen2009exact}. Enumerations of all motifs having variable lengths has also been researched~\cite{10.1109/ICDM.2013.27,yasser2014}. In a streaming scenario an algorithm can not rely on accessing the full past series, therefore we need to find the top-k motif search via an on-line method as in~\cite{doi:10.1137/1.9781611972818.86}. In addition, the statistical significance of the motifs found has also been a topic of interest~\cite{MotifsStat,Castro:2012:SMT:2305476.2305481}.

\emph{Note:} Finding motif-pairs is equivalent to the problem of locating the closest pair of points in a geometrical space and is a historic problem in computational geometry~\cite{Cormen:2001:IA:580470}.

\subsection{Motif Discovery}

Repeating patterns in sequential data have initially been studied in bio-informatics~\cite{Buhler:2001:FMU:369133.369172}. However, finding motifs is beneficial in understanding physiological human data~\cite{Syed:2010:MDP:1644873.1644875}, while being also useful in understanding behavioral patterns of living organisms~\cite{brown2013dictionary}. The concept of recurrent patterns was transferred to the realm of time-series data under the term "motifs" \cite{Patel:2002:MMM:844380.844710} and a search-based approach to discovering motifs was proposed. In order to find motifs that are immune to noisy variations, a probabilistic search of time-series motifs was based on random projections~\cite{Chiu:2003:PDT:956750.956808}. Another work has explored the employment of uniform scaling as the similarity distance used for discovering the motifs~\cite{Yankov:2007:DTS:1281192.1281282}. Furthermore, a hybrid combination of supervised and unsupervised learning has been used for searching recurring patterns~\cite{1183920}. The first step involves a \emph{teacher} which labels whether or not a time series includes a particular pattern, while in the next step an unsupervised learning from the series in order to reconstruct the \emph{teacher} is exploited. The task of finding the most recurring motifs has also been tackled through searching for candidate motifs organized in a tree structure~\cite{tubao2005}. 

The brute-force approach which tries out every segment (sub-sequence) as a potential motif has a quadratic complexity in the number of segments. Therefore approximate motif discovery methods have been exploited. Conversion of motifs into a symbolic representation (named SAX) is a pre-processing alternative~\cite{todorovski2006}. Over the new representation an agglomerative clustering can be used to find motifs~\cite{todorovski2006}. A scalable alternative that can approximately discover multi-resolution motifs in a single scan utilizes different cardinalities of the symbolic representation~\cite{MrMotif}. Last but not least, a scalable version of the pair-wise motifs has been extended to the general motifs discovery for large-scale data~\cite{Mueen:2009:FTS:1674659.1677089}.

Given the widespread of multi-dimensional time series, there has also been interest in mining multi-dimensional motifs too. Several strategies were inspected, where motifs span all versus a subset of the dimensions, with or without temporal overlap~\cite{Minnen:2007:DSM:1441428.1442125}. The algorithm is based on random projections of the symbolic sub-sequence representations \cite{Minnen:2007:DSM:1441428.1442125}. Discovering regions of high density in the space of sub-sequencies is another alternative to mining multivariate motifs~\cite{Minnen:2007:DMM:1619645.1619744}. Graph clustering implemented as a two-staged algorithm was also employed in detecting multidimensional motifs~\cite{Vahdatpour:2009:TUA:1661445.1661647}. In the first step single-dimensional motifs are discovered and later blended through clustering~\cite{Vahdatpour:2009:TUA:1661445.1661647}.
 
Since motifs are previously unknown patterns, there is little information on the motifs' lengths too. Under such a reality authors attempted to discover the optimal motif length, for instance by inspecting the compressibility of the data~\cite{6729632}. In addition, variable-length motifs can be extracted using a grammar-inspired inference process~\cite{Li:2010:AVT:1814245.1814255}. Interest has been attracted in terms of visualizing variable-length motifs~\cite{DBLP:conf/sdm/LiLO12}, finding them in linear time~\cite{catalano2006}, or using them for classification purposes~\cite{yin2014}.																													

\textbf{In contrast to the related work}, our novel contribution relies in computing an optimal set of motifs given a threshold distance and the motifs' length. We are the first to propose a principled optimization method for the task. As a consequence, our approach leads to significantly improved motif quality (frequency) compared to brute-force search.

\section{Preliminaries}

\subsection{Notations}

\subsubsection{Time Series and Motifs}

A time series is a long ordered sequence of real-valued measurements. Such a series is abstracted as a list of $J$-many Z-normalized sliding-window segments of length $L$ and is denoted as $S \in \R^{J \times L}$. On the other hand, a repetitive pattern, a.k.a motif, is simply a sequence of $L$ points. The definition can be generalized to a set of $K$-motifs and consecutively denoted as $M \in \R^{K \times L}$.



\subsubsection{Motif Frequency}

The occurrence frequency of a motif is defined as the \emph{nontrivial} (see Section~\ref{sec:nontrivialMatches}) number of matches between a motif and all the normalized segments of the time series. The current approach of counting the matching frequency of the $k$-th motif, denoted $M_{k,:} \in \R^{L}$, iterates over all the $j \in \{1,\dots,J\}$ sliding window segments $S_{j,:}$ and check whether the motif of interest matches the segments within a \textbf{\emph{threshold distance}} $T \in \R^{+}$. 

\vspace{-0.4cm} 

\begin{eqnarray}
\label{eq:hardFrequencyEquation} 
\mathcal{F}(M) &=& \sum_{k=1}^{K} \sum_{j=1}^{J} \mathcal{F}_{k,j} \\
\label{eq:hardFrequencyPerSegmentEquation} 
\mathcal{F}_{k,j} &=& \begin{cases}
1 & \text{if } \left( \sum_{l=1}^{L} \left(M_{k,l} - S_{j,l}\right)^2 \right) < T  \\
0 & \text{otherwise}
\end{cases}
\end{eqnarray}

Equation~\ref{eq:hardFrequencyEquation} presents the formalism for the overall frequency as a sum of motifs' frequencies, while Equation~\ref{eq:hardFrequencyPerSegmentEquation} encapsulates the concept of a \emph{match}. If the distance between a segment $S_{j,:}$ and a motif $M_{k,:}$ is less than the threshold $T$, then a matching value of one is granted.

\subsection{Problem Definition}

\subsubsection{Optimal Motifs}

Following the established literature definition, the only optimality criterion of a motif is its frequency at a particular distance threshold. Therefore, the only legitimate metric to compare the qualities of motifs is frequency (a.k.a. support, or number of matches). The optimal motifs $M^{*}$ for a time series are defined in Equation~\ref{eq:optimalMotifs} as the candidate motifs $M$ that achieve the maximum frequency value $\mathcal{F}(M)$ from Equation~\ref{eq:hardFrequencyEquation}. There is, nevertheless, an important constraint in the search for motifs: The $K$ motifs should be different from each other \cite{Patel:2002:MMM:844380.844710}, otherwise, the motifs risk being close variations of the single most repetitive motif. Such a constraint is presented under a "such that (s.t.)" clause in Equation~\ref{eq:optimalMotifs}, which enforces each pair of motifs $(M_{k,:}, M_{p,:})$ to be different from each other by a distance of at least $2T$ (so each pair does not overlap within a threshold $T$, details in \cite{Patel:2002:MMM:844380.844710}).

\vspace{-0.3cm} 

\begin{eqnarray}
\label{eq:optimalMotifs} 
M^{*} &:=& \argmax_{M \in \R^{K \times L}} \;\;\; \mathcal{F}(M) \\ 
\nonumber
& \text{s.t.:} & \left( \sum_{l=1}^{L} \left( M_{k,l} - M_{p,l} \right)^2 \right) > 2T, \\ \nonumber 
&& \forall k \in \{1,\dots,K\}, \forall p \in \{k+1,\dots,K\}
\end{eqnarray}

\subsubsection{Trivial Matches}
\label{sec:nontrivialMatches}

Stated shortly, trivial matches are consecutive segments which match the same motif~\cite{Patel:2002:MMM:844380.844710}. For instance, this case might happen if the sliding window is incremented by one. In that case two subsequent segments will share exactly $L-1$ points and therefore the distance of any motif to those close-by segments will be very similar. Some related work increment the sliding window by an offset of points, therefore trivial matches can be trans-passed at the risk of potentially missing certain matches~\cite{MrMotif,Minnen:2007:DMM:1619645.1619744}. However, in our paper all the reported figures on frequency do not include any trivial match throughout the experiments. 

\subsection{Searching The Motifs}

The state-of-the-art methods referred in Section~\ref{sec:relatedWork} focusing on searching motifs are primarily concerned with trying candidate motifs from the series segments. Despite proposing important novelties in their scope (scalability, length analysis, etc \dots) still these techniques are upper bounded in terms of quality by the brute-force motif search. 

\begin{algorithm}[h]
   \caption{BruteForceMotifSearch()}
   \label{alg:bruteForceSearchMotifs}
\begin{algorithmic}[1]
\STATE {\bfseries Input: } Threshold $T \in \R^{+}$, Motif length $L \in \N^{+}$, Number of Motifs $K \in \N^{+}$, Segments $S \in \R^{J \times L}$  
\STATE {\bfseries Output: } $M \in \R^{K \times L}$
\STATE $\text{// Precompute frequencies of all segments}$ 
\FOR{$j=1,\dots,J$}
\STATE $\mathcal{F}_j \leftarrow 0 $
\STATE $\text{lastMatchIndex} \leftarrow - \infty$
\FOR{$r=1,\dots,J$}
\IF{$ ||S_{j,:} - S_{r,:}||_2^2 < T$}
	\STATE $\text{// Avoid trivial matches}$ 
	\IF{$r - \text{lastMatchIndex} > 1$}
		\STATE $\mathcal{F}_j \leftarrow \mathcal{F}_j + 1 $
	\ENDIF 
	\STATE $\text{lastMatchIndex} \leftarrow r$ 
\ENDIF
\ENDFOR
\ENDFOR
\STATE $\text{// Select top-K motifs}$ 
\FOR{$k=1,\dots,K$}
    \STATE $\text{best}_j \leftarrow 0$ 
	\FOR{$j=1,\dots,J$} 
		\STATE $\text{// Check if the j-th segment is diverse}$
		\IF{$ ||S_{j,:} - M_{p,:}||_2^2 > 2T, \; \forall p=k-1,\dots,1$}
			\IF{$\mathcal{F}_{\text{best}_j} > \mathcal{F}_{j}$}
		       \STATE $\text{best}_j \leftarrow j$ 
	        \ENDIF
       \ENDIF
	\ENDFOR    
	\STATE $M_{k,:} \leftarrow S_{\text{best}_j,:}$   
\ENDFOR
\STATE {\bfseries return } $M$
\end{algorithmic}
\end{algorithm}

Algorithm~\ref{alg:bruteForceSearchMotifs} describes a speed-wise naive, yet qualitatively search-optimal implementation of a brute-force motif search. We can pre-compute the frequencies of all series segments in $\mathcal{O}(J^2 L)$ runtime complexity and then search the top-K motifs using the computed frequencies in $\mathcal{O}(K^2 J L)$ time. Since $K$ is typically a small number compared to the segments $J >> K$, therefore the overall brute-force search has a complexity of $\mathcal{O}(J^2 L + K^2 J L) \sim \mathcal{O}(J^2 L)$, meaning quadratic in the number of segments. In this paper we propose a learning (not searching) method that outputs motifs having higher frequencies than those discovered by the brute-force approach.

\section{Proposed Method}

\subsection{Motivation}
\label{sec:motivation}

\begin{figure}[h]
\centering
\includegraphics[scale=0.5, trim=1cm 14cm 2.0cm 7cm]{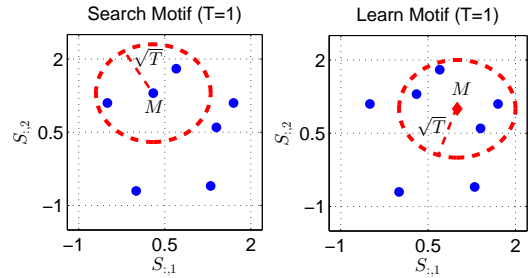}
\caption{Motif found by searching (left) yields 3 matches while learning a latent motif (right) yields 4 matches}
\label{fig:searchVsLearn}
\end{figure}

The state-of-the-art methods used for finding motifs are based on \emph{searching} for the most frequently occurring candidate \emph{segment}. In other words, any motif has to explicitly occur as a series segments $M_{k,:} \in S, \; \forall k \in \N_{k=1}^K$. Unfortunately, such constrained motifs are very restricted in the finite space of possible values they can have, compared to the space of real matrices $M \in \R^{K \times L}$ (infinitely more candidates than $M \in S$). In this paper, we hypothesize and empirically show that the optimal motifs are located in the space of real numbers $M \in \R^{K \times L}$, while the space of segments contains sub-optimal motifs. Figure~\ref{fig:searchVsLearn} provides a hint for the comparison between restricted motifs $\left( M \in S \right)$ and un-restricted optimal ones. From a geometrical perspective the segments and the motifs are points in an $L$-dimensional space. In the example of Figure~\ref{fig:searchVsLearn} the segments and motifs have a length of 2, thus the scenario is 2-dimensional.

The frequency of a motif $M$, given a threshold $T$, can be interpreted as the number of segment points (blue in the illustration) that lies within a radius of the threshold distance from the motif (shown in red). The radius is $\sqrt{T}$ because we used the squared-Euclidean distance in Equation~\ref{eq:hardFrequencyPerSegmentEquation}, however this poses no problems since $T$ is anyway a hyper-parameter of our method. The most frequently occurring motif is defined to be the point that covers the maximum number of blue points (segments) inside the circle of radius $\sqrt{T}$ that is centered at the motif, hence the densest geometrical ball~\cite{tubao2005}. The best segment-motif is shown in the left plot of Figure~\ref{fig:searchVsLearn} and has a frequency of three. However, the optimal motif is located in the right plot and has a frequency of four. As clearly seen, the optimal solution is \emph{hidden} in the space of real numbers, outside the very restricted set of segment points. The method proposed in this paper \emph{learns} optimal motifs lying in the real-numbers space through a tailored numerical optimization technique. Even though the aforementioned 2-dimensional example was \emph{created} to awake the reader on the need for \emph{learning} motifs, still empirical results of Section~\ref{sec:empiricalResults} will demonstrate that learning motifs yields more frequently occurring patterns, compared to searching them, on real-life time series.

\subsection{Smooth (Differentiable) Motif Frequency}

We are going to find the optimal motif through a mathematical maximization of the frequency as a function of the motifs. Unfortunately, the frequency of Equation~\ref{eq:hardFrequencyEquation} has two problems (i) it is not continuous at point $||M_{k,:}-S_{j,:}||=T$ and (ii) first derivative is zero in all other points (i.e. frequency is flat having values 1 or 0). Therefore we cannot compute the optimal motifs using gradient-based optimization. However, we can use a differentiable approximation for the frequency function using the Gaussian kernel of Equations~\ref{eq:softFrequency}-\ref{eq:softFrequencyPerSegment}.

\vspace{-0.3cm} 

\begin{eqnarray}
\label{eq:softFrequency}
\hat{\mathcal{F}}(M) &=& \frac{1}{KJ} \sum_{k=1}^{K} \sum_{j=1}^{J} \hat{\mathcal{F}}_{k,j} \\
\label{eq:softFrequencyPerSegment}
\hat{\mathcal{F}}_{k,j} &=&  e^{- \frac{\alpha}{T} \sum_{l=1}^{L} \left(M_{k,l} - S_{j,l}\right)^2 }
\end{eqnarray}

\begin{figure}[h]
\centering
\includegraphics[scale=0.48, trim=2cm 14cm 2.0cm 6cm]{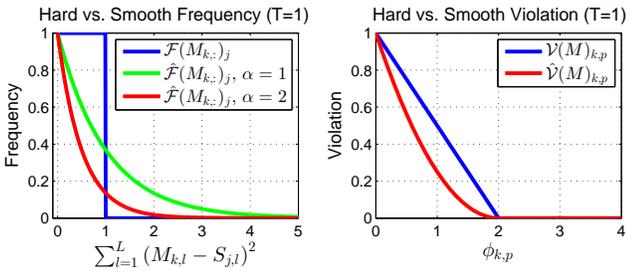}
\caption{Smooth vs. Hard Variants of Frequency (left) and Diversity Violation (right)}
\label{fig:exampleFrequencyViolation}
\end{figure}

The smooth frequency function of Equation~\ref{eq:softFrequencyPerSegment} is both an accurate approximation to the frequency measure from Equation~\ref{eq:hardFrequencyPerSegmentEquation}, but also a differentiable alternative, as illustrated in Figure~\ref{fig:exampleFrequencyViolation} (left plot). The parameter $\alpha$ controls the smoothness of the soft frequency. For optimization reasons (details in Section~\ref{sec:motifLearningOptimization}) the frequency sum of Equation~\ref{eq:softFrequency} is divided by $KJ$ to limit the value of $\hat{\mathcal{F}}$ between 0 and 1. In terms of notation, the approximated frequency is distinguished by a hat ($\mathcal{F}$ vs $\hat{\mathcal{F}}$).

\subsection{Motif Diversity Violation} 

As previously described in Equation~\ref{eq:optimalMotifs} the motifs need to be distant by a margin of $2T$. We call such a property as \emph{motif diversity}. In that line, this section is devoted to formalizing a differentiable penalty function for the violations of the distances among motifs from the diversity threshold of $2T$. As a first step, the distance between two motifs $M_{k,:} \in \R^L$ and $M_{p,:} \in \R^L$ is defined as $\phi_{k,p}: \left(\R^L \times \R^L\right) \rightarrow \R$ and formalized in Equation~\ref{eq:distanceBetweenMotifs}.

\vspace{-0.2cm} 
\begin{equation}
\label{eq:distanceBetweenMotifs}
\phi_{k,p} = \sum_{l=1}^{L} \left( M_{k,l} - M_{p,l} \right)^2 
\end{equation}

The distance $\phi_{k,p}$ of any pair of motifs $M_{k,:}, M_{p,:}$ should obey to the diversity constraint shown in Equation~\ref{eq:motifDiversityConstraint}.

\vspace{-0.2cm} 
\begin{eqnarray}
\label{eq:motifDiversityConstraint}
\phi_{k,p} > 2T, \;\;\; \forall k \in (\{1,\dots,K\}, \; \forall p \in \{k+1,\dots,K\})
\end{eqnarray}

We introduce the concept of \emph{diversity violation} by Equations~\ref{eq:hardViolationPenalty}-\ref{eq:hardViolationPenaltyPerPair}. For each of the $\frac{K(K-1)}{2}$ pairs of motifs, the violation is 0 if the distance between the pair motifs is greater than $2T$. Otherwise, if the distance is zero then the motifs are identical (hence not at all diverse) and a maximum violation of one is returned. For all the distances between $0$ and $2T$ a linear violation between 0 and 1 is returned as formalized in Equation~\ref{eq:hardViolationPenaltyPerPair}. The constant term $\frac{2}{K(K-1)}$ makes sure that the violation function has a range between 0 and 1, the same range as the approximative frequency. 

\vspace{-0.4cm} 
\begin{eqnarray}
\label{eq:hardViolationPenalty}
\mathcal{V}(M) &=& \frac{2}{K(K-1)} \sum_{k=1}^{K} \sum_{p=k+1}^{K} \mathcal{V}_{k,p} \\ 
\label{eq:hardViolationPenaltyPerPair}
\mathcal{V}_{k,p} &=& \begin{cases}
1 - \frac{ \phi_{k,p}}{2T} & \phi_{k,p} < 2T \\
0 & \phi_{k,p} \ge 2T
\end{cases}  
\end{eqnarray}

Despite achieving its aim, the violation penalty of Equations~\ref{eq:hardViolationPenalty}-\ref{eq:hardViolationPenaltyPerPair} still it suffers in terms of differentiability at the point $\phi_{k,p}=2T$. Therefore, we are proposing a smooth and differentiable variant of the violation penalty in Equations~\ref{eq:violationPenalty}-\ref{eq:violationPenaltyPerSegment} by squaring the hard violation of Equation~\ref{eq:hardViolationPenaltyPerPair}.

\vspace{-0.4cm} 
\begin{eqnarray}
\label{eq:violationPenalty}
\hat{\mathcal{V}}(M) &=& \frac{2}{K(K-1)} \sum_{k=1}^{K} \sum_{p=k+1}^{K} \hat{\mathcal{V}}_{k,p} \\ 
\label{eq:violationPenaltyPerSegment}
\hat{\mathcal{V}}_{k,p} &=&  \begin{cases}
\left( 1 - \frac{\phi_{k,p}}{2T} \right)^2   & \phi_{k,p} < 2T \\
0 & \phi_{k,p} \ge 2T
\end{cases}  
\end{eqnarray}

As in the case of the frequency, we denote the smooth approximative version of the violation penalty by a hat ($\mathcal{V}$ for hard and $\hat{\mathcal{V}}$ for smooth). The violation penalty as a function of the distance between motif pairs is depicted in the right plot of Figure~\ref{fig:exampleFrequencyViolation}.

\subsection{Motif Learning Through Optimization}
\label{sec:motifLearningOptimization}

This section fuses the smooth motif frequency and smooth motif diversity violation into a meaningful objective function. Our aim is to learn a set of $K$ motifs that \emph{maximize} the frequencies and \emph{minimize} (\emph{have no}) violations. Such an objective can be elegantly constructed as the maximization task of Equation~\ref{eq:objectiveFunctionNovelty}. 

\vspace{-0.3cm} 
\begin{eqnarray}
\nonumber
M^{*} &=& \argmax_{M} \;\; \mathcal{O}(M) \\
\label{eq:objectiveFunctionNovelty}
&=& \argmax_{M} \;\; \hat{\mathcal{F}}(M) - \hat{\mathcal{V}}(M)
\end{eqnarray}

The universally optimal motifs are those which achieve the universal maximum value of our objective function $O(M) = \hat{\mathcal{F}}(M) - \hat{\mathcal{V}}(M)$. As both terms are positive, the objective is maximized for the highest motif frequencies and zero violations. In this paper we will optimize the objective function through gradient ascent motif updates in a series of iterations. Since both ranges of $\hat{\mathcal{F}}$ and $\hat{\mathcal{V}}$ are between 0 and 1, no term over-scales the other and the overall learning does converge. In our preliminary experiments we found out that a trade-off coefficient $\beta$ in the form $\hat{\mathcal{F}}(M) - \beta \hat{\mathcal{V}}(M)$ was not needed as both terms converge quickly.

\begin{figure}[h]
\centering 
\includegraphics[scale=0.48, trim=1.7cm 9.2cm 1.5cm 7.2cm]{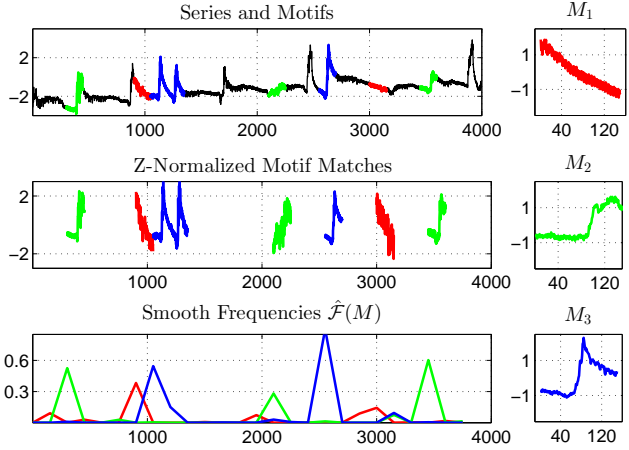} 
\caption{Top-3 motifs from the "Insect B" time series $(L=150, T=61, \eta=0.3, I=300, \alpha=2)$}  
\label{fig:topMotifsIllustration}
\end{figure} 

The output of the learning process is a set of motifs $M$, as shown in Figure~\ref{fig:topMotifsIllustration} for the "Insect B" time series. In this illustration the top three motifs $(K=3)$ are shown on the right plots, while the matches of the motifs on the time series are shown in the upper-left plot. Z-normalized versions of the matched segments are shown in the middle-left plot and the lower-left plot illustrates the per-segment smooth frequency scores of the motifs. 

\subsubsection{Gradient Ascent Optimization}
\label{sec:gradientAscentOptimization}

Since the objective function of Equation~\ref{eq:objectiveFunctionNovelty} is a subtraction of frequency and diversity violations, the partial gradient of the objective function with respect to each point $l$ of any $k$-th motif is decomposable as shown in Equation~\ref{eq:ojectiveDerivative}.

\vspace{-0.2cm} 

\begin{eqnarray}
\label{eq:ojectiveDerivative}
\frac{\partial \mathcal{O}(M) }{ \partial M_{k,l} } &=& \frac{\partial \hat{\mathcal{F}}(M) }{ \partial M_{k,l} } - \frac{\partial \hat{\mathcal{V}}(M) }{\partial M_{k,l}} 
\end{eqnarray}

The partial derivative of the smooth frequency with respect to the motif is computed as the first derivative of Equation~\ref{eq:softFrequency} in terms of $M$ and shown below in Equation~\ref{eq:derivativeFwrtM}.

\begin{eqnarray}
\label{eq:derivativeFwrtM}
\frac{\partial \hat{\mathcal{F}}(M) }{ \partial M_{k,l} } &=& \frac{-2 \alpha}{K J T} \sum_{j=1}^{J}  \left(M_{k,l} - S_{j,l}\right) \hat{\mathcal{F}}_{k,j} 
\end{eqnarray}

Similarly the partial derivative of the diversity violation with respect to each motif's point is defined in Equation~\ref{eq:derivativeVwrtM}.

\begin{eqnarray}
\label{eq:derivativeVwrtM}
\frac{\partial \hat{\mathcal{V}}(M)}{\partial M_{k,l}} &=& \frac{2}{K(K-1)} \sum_{q=1}^{K} \frac{\partial \hat{ \mathcal{V}}_{k,q}}{\partial M_{k,l}} \\ \nonumber
\frac{\partial \hat{\mathcal{V}}_{k,q}}{\partial M_{k,l}} &=& \begin{cases}
  \frac{\left(\phi_{k,q} - 2T \right) \left(M_{k,l}-M_{q,l} \right)}{T^2}   & \phi_{k,q} < 2T \\ 
0 & \phi_{k,q} \ge 2T
\end{cases} 
\end{eqnarray}

\newpage

\subsection{Learning Algorithm}

Having defined the partial derivative needed for gradient ascent, we can present the complete learning method. Our method is detailed in Algorithm~\ref{alg:learnMultipleMotifs} and in this section we will explain the steps of the algorithm in detail. There are a set of hyper-parameters to the learning process, starting with the frequency smoothness $\alpha$. The other important hyper-parameters are the number of motifs $K$, the threshold $T$ and the motif length $L$, to be set by a practitioner. The learning rate $\eta$ and the number of iterations $I$ are less critical hyper-parameters that control the number of steps needed until convergence. For small learning rates and large number of iterations, the convergence is safely achievable.

\begin{figure}[h]
\centering 
\includegraphics[scale=0.5, trim=2cm 10.5cm 1.5cm 6cm]{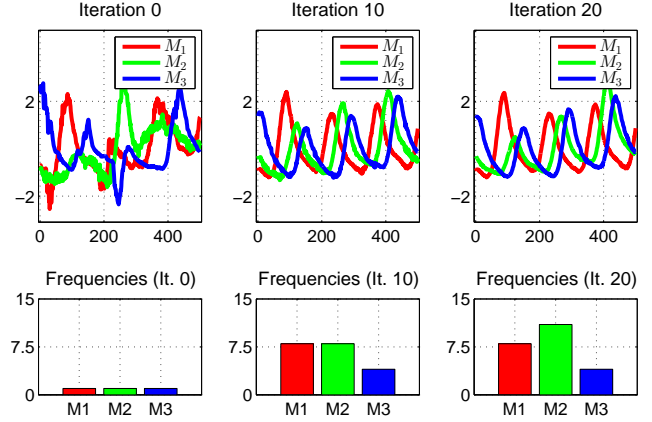} 
\caption{Metamorphosis of three motifs on the "EOG" time series $(L=150, T=58, \eta=0.3, I=300, \alpha=2)$} 
\label{fig:motifMetamorphosis}
\end{figure} 

\begin{algorithm}[h]
   \caption{LearnMotifs()}
   \label{alg:learnMultipleMotifs}
\begin{algorithmic}[1]
   \STATE {\bfseries Input:} Threshold $T \in \R^{+}$, Motif length $L \in \N^{+}$, Number of Motifs $K \in \N^{+}$, Segments $S \in \R^{J \times L}$, Learning Rate $\eta \in \R^{+}$, Number of iterations $I \in \N^{+}$,  Smoothness $\alpha \in \R^+$  
   \STATE {\bfseries Output:} Motif $M \in \R^{K \times L}$ 
   \STATE \text{// Initialize random motifs and gradient accumulators:}
   \STATE $M \leftarrow {\left(S_{\mathcal{U}(1,J),:}\right)}^K, \nabla \leftarrow 0^{K \times L}$
   \STATE \text{// Initialize constant values:}
   \STATE $c_{\hat{\mathcal{V}}} \leftarrow \frac{2}{K(K-1)T^2}, \;\; c_{\hat{\mathcal{F}}} \leftarrow \frac{-2 \alpha}{K J T} $
   \STATE \text{// Iterate the learning method:}
   \FOR{iter$=1,\dots,I$}
   \STATE \text{// Precompute the per-segment occurence scores:}
   \STATE $\hat{\mathcal{F}}_{k,j} \leftarrow e^{- \frac{\alpha}{T} \sum_{l=1}^{L} \left(M_{k,l} - S_{j,l}\right)^2 } \;\;\; \forall k \in \N_{1}^{K}, \forall j \in \N_{1}^{J}$
   \STATE \text{// Precompute the pair-wise motif distances:}
   \STATE $\phi_{k,q} \leftarrow \sum\limits_{l=1}^{L} \left(M_{k,l}-M_{q,l}\right)^2, \;\;\; \forall k \in \N_{1}^{K}, \forall q \in \N_{1}^{K}$
   \STATE\text{// Update the motifs :} 
   \FOR{$k=1,\dots, K; \;\; l=1,\dots,L$} 
   \STATE \text{// Gradient of frequency w.r.t. the motif:}
    \STATE $\frac{\partial \hat{\mathcal{F}}(M) }{ \partial M_{k,l} } = c_{\hat{\mathcal{F}}} \sum\limits_{j=1}^{J}  \left(M_{k,l} - S_{j,l}\right) \hat{\mathcal{F}}_{k,j}$
    \STATE \text{// Gradient of diversity violation w.r.t. the motif:}
    \STATE $\frac{\partial \hat{\mathcal{V}}(M)}{\partial M_{k,l}} = c_{\hat{\mathcal{V}}} \sum\limits_{q=1}^{K} \begin{cases}
     \left(\phi_{k,q} - 2T \right) \left(M_{k,l}-M_{q,l} \right)   & \phi_{k,q} < 2T \\ 
    0 & \phi_{k,q} \ge 2T
    \end{cases} $
    \STATE \text{// Gradient of the final objective w.r.t. the motif:}
    \STATE $\frac{\partial \mathcal{O}(M)}{\partial M_{k,l}} \leftarrow \frac{\partial \hat{\mathcal{F}}(M)}{\partial M_{k,l}} - \frac{\partial \hat{\mathcal{V}}(M)}{\partial M_{k,l}} $ 
    \STATE \text{// Update the history of gradients:}
    \STATE $\nabla_{k,l} \leftarrow \nabla_{k,l} + \left( \frac{\partial \mathcal{O}(M)}{\partial M_{k,l}} \right)^2$ 
    \STATE\text{// Update the motif point:} 
   \STATE $M_{k,l} \leftarrow M_{k,l} + \frac{\eta}{ \sqrt{ \nabla_{k,l} } }  \frac{\partial \mathcal{O}(M)}{\partial M_{k,l}}$
    \ENDFOR
   \ENDFOR
   \STATE {\bfseries return } $M$
\end{algorithmic}
\end{algorithm}

The algorithm starts with a set of motifs initialized from random segments and updates them in the direction of the partial gradients using a learning rate step size. The learning rate is dynamically updated per each point of each motif using an adaptive technique known as AdaGrad \cite{Duchi:2011:ASM:1953048.2021068}. We accumulate the square of the partial gradients into accumulators denoted by $\nabla$. In order to speed-up the updates we pre-compute the per-segment frequencies $\hat{\mathcal{F}}_{k,j}$ and pair distances $\phi_{k,q}$ in lines 9-12. Then every point of each motif $M_{k,l}$ is updated in the positive direction of the derivative in lines 13-25. The partial gradients correspond to the ones previously explained in Section~\ref{sec:gradientAscentOptimization}. The update of line 24 adjusts the learning rate by the square root of the accumulated square gradients~\cite{Duchi:2011:ASM:1953048.2021068}. 

As a consequence of the gradient ascent updates, the motifs undergo a \emph{metamorphosis} as is shown in Figure~\ref{fig:motifMetamorphosis} for the "Full EOG" time series. The illustrative motifs are learned on the first 10000 non-overlapping segments of the time series having length $L=150$. At the beginning (Iteration 0) the motifs are random and the corresponding frequencies zero, however the motifs start to take form after approximately 20 iterations and converge after 40 iterations. The metamorphosis of the motifs is conducted such that their matching frequencies (lower plots) are maximized. 

\begin{figure}[h!] 
\centering 
\includegraphics[scale=0.45, trim=2cm 14cm 2.0cm 6cm]{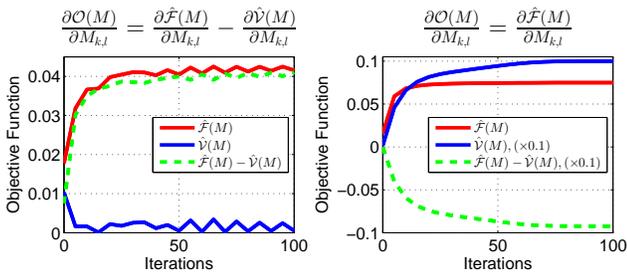}
\caption{Convergence on "Insect B" dataset ($K=5, T=382, \eta=0.3$) }
\label{fig:convergenceExample} 
\end{figure} 

\subsection{Convergence of The Learning Algorithm}

The learning algorithm converges by updating the motifs so that the approximative frequency is maximized and the diversity violations minimized to zero as shown in Figure~\ref{fig:convergenceExample} (left plot) for an execution on the "Insect B" dataset. It is worth noting that the inclusion of the penalty on the diversity violation is crucial for preserving the diversity constraint. An experiment is shown on the right plot of Figure~\ref{fig:convergenceExample}. In this experiment the line 24 of Algorithm~\ref{alg:learnMultipleMotifs} is edited so the motifs are updated only with respect to the frequency and not diversity violation (see plot title). As we can clearly see, maximizing the frequencies without penalizing diversity violations causes the motifs to be similar to each other. That is demonstrated by the fact that the violation measure increases, as shown in the right plot of Figure~\ref{fig:convergenceExample}.

\section{Optimality of Our Method}

\begin{figure}[h]
\centering
\includegraphics[scale=0.7, trim=2.1cm 16.8cm 2.0cm 6.9cm]{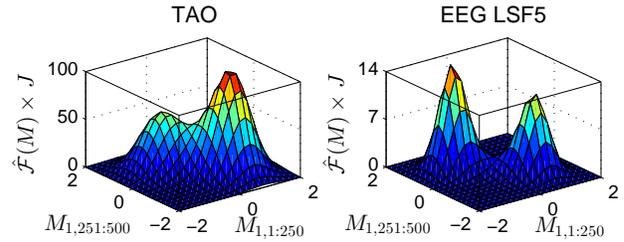}
\caption{Non-concave frequency $\hat{\mathcal{F}}(M)$ as a function of motif values $M_{1,:}$ on TAO and EEG LSF5 time-series datasets, Parameters: $L=500, T=100, \alpha=2$}
\label{fig:nonConcaveFrequency}
\end{figure}

The objective function of Equation~\ref{eq:objectiveFunctionNovelty} is not concave, because the frequency function is a sum of Gaussians and not concave. We demonstrate the non-concavity of the frequency function in Figure~\ref{fig:nonConcaveFrequency} using the TAO and EEG LSF5 datasets. Here we generate all possible motifs of length 500 using two values, (for the sake of a 3d-plot), one value for all the first 250 points in X-axis and another value for the last 250 points in the Y-axis. As can be clearly seen, frequency is not a concave function in terms of motifs and has multiple local maxima.

In case of non-concave functions (or non-convex for minimization problems), an effective cook-book solution is to combine gradient descent with a \emph{random-restart} strategy~\cite{lones2011sean}. In order to avoid getting stuck in local maxima, the gradient descent optimization is restarted multiple times with random initial values for the motifs. The run that achieves the highest $\mathcal{F}(M)$ is selected, as is formalized in Equation~\ref{eq:randomRestarts}, where the number of restarts is denoted by $R \in \N$. It is important to recognize that we select the motifs yielding the highest hard frequency $\mathcal{F}$, not the proxy smooth one $\hat{\mathcal{F}}$. The hard frequency $\mathcal{F}$ does \textbf{avoid} counting \textbf{trivial} matches in our implementation.

\vspace{-0.4cm}
\begin{eqnarray}
\label{eq:randomRestarts} 
M^{*} &:=& \argmax_{M^{(r)}, \; r={1,\dots,R}} \mathcal{F}(M^{(r)}) \\
\nonumber 
&s.t.& M^{(r)} \leftarrow \text{LearnMotif() from Alg~\ref{alg:learnMultipleMotifs}}
\end{eqnarray}

\begin{figure}[h]
\centering
\includegraphics[scale=0.47, trim=0.8cm 14.8cm 2.0cm 6cm]{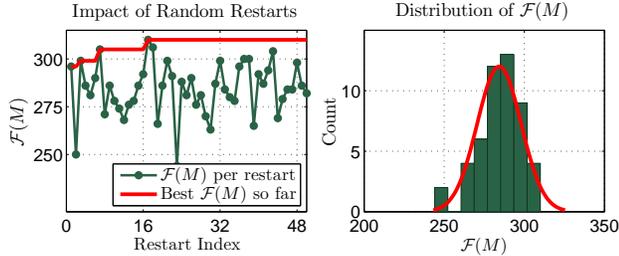}
\caption{Impact of Random Restarts on $\mathcal{F}(M)$; "TAO" time-series dataset with hyper-parameters $L=500, K=10, \eta=0.3, I=300, T=109.6$}
\label{fig:randomRestarts}
\end{figure}

Figure~\ref{fig:randomRestarts} illustrates the effect of 50 random restarts on the frequency function $\mathcal{F}(M)$ values over the TAO dataset. On the left plot we see that the maximum values of the objective are reached after a few restarts. The distribution of the frequency values, shown in the right plot, demonstrates that the histogram is normally distributed. That means there is a normal probability that a restart will yield an optimal value on the right portion (maximal) of the values within.

\subsection{Runtime Algorithmic Complexity}

The runtime complexity of Algorithm~\ref{alg:learnMultipleMotifs} is determined by the pre-computation steps and the update steps. Computing of the frequency terms has an algorithmic complexity order of $\mathcal{O}(R I K J L)$, while computing the pairwise distances has a computational complexity of $\mathcal{O}(R I K^2 L)$. The computation of the partial gradients of the frequency with respect to the motifs has a complexity of $\mathcal{O}(R I K J L )$. Similarly the complexity of computing the gradients of the diversity violation with respect to the motif has a complexity of $\mathcal{O}(R I K^2 L)$. The overall complexity of the algorithm is $\mathcal{O}(RIKJL + RIK^2L + RIKJL + RIK^2L)$, which translates to $\mathcal{O}(2 R I K \left(J+K\right)L) \sim \mathcal{O}(R I K J L)$ since $K << J$. The brute force search on the other hand, has a complexity of $\mathcal{O}(J^2 L)$ which is quadratic in terms of the number of segments $J$. In contrast our method is linear in terms of the number of segments $J$ and faster than the brute-force search in case $RIK < J$. It is worth reminding that our algorithm learns optimal motifs (brute-force finds non-optimal motifs) and the primary strength is quality at a feasible runtime.

\section{Empirical Results}
\label{sec:results}

\subsection{Experimental Setup}

We compare the quality of the proposed methods against the brute-force search strategy using a battery of six time-series datasets from diverse application domains. In addition, we employ an evaluation protocol which compares the frequencies of the computed motifs per different number of motifs, motif lengths and distance thresholds.

\subsubsection{Datasets}

\begin{itemize}
\item \textbf{Insect B} is a time series of insect behavior data and has a length of 73929 points~\cite{mueen2009exact}. 
\item \textbf{TAO} is  a long time series representing Tropical Atmosphere Ocean temperature measurements having 741528 measurements\footnote{\url{www.pmel.noaa.gov/tao}}.
\item \textbf{RandomWalk} is a time-series dataset consisting of 1000000 points, among which motifs at randomly selected time-stamps are implanted \cite{mueen2009exact}.
\item \textbf{EEG} is a series of 1802136 continuous measurements from electroencephalographic sensors, measuring voltage differences across the scalp \cite{mueen2009exact}.
\item \textbf{Salinity} is a time series containing recordings on the level of oceanic salt concentration. The data has a length of 2324134 points and is provided by the National Oceanographic Data Center\footnote{\url{http://www.nodc.noaa.gov/General/salinity.html}}.
\item \textbf{EOG} is the longest series in our collection consisting of 8099500 points. The data is collected by an Electro-Oculogram and represent electrical potential between the front and the back of a human eye~\cite{PhysioNet}.
\end{itemize} 

\subsubsection{Baseline}

Many motif discovery method are based on searching for frequent patterns among the series segments (e.g. \cite{Patel:2002:MMM:844380.844710,Yankov:2007:DTS:1281192.1281282,Chiu:2003:PDT:956750.956808,DBLP:conf/sdm/LiLO12,Li:2010:AVT:1814245.1814255}, enumerated in a broader scope in Section~\ref{sec:relatedWork}). While those search-based methods are successful in terms of scalability, data representation, on-line learning, etc..., they are still upper bounded in quality (a.k.a. frequency) by the Brute-Force search. That is trivial to show, because all the frequent sub-sequences those methods could find are also detectable by Brute-Force search. In that aspect, it is sufficient to demonstrate that our method is superior to Brute-Force searching in terms of \textbf{quality} (a.k.a. \textbf{frequency}) and that naturally translates into qualitative superiority against all the other scalable/approximate/on-line search-based methods. 

\subsubsection{Evaluation Protocol}

We will compare against the brute-force search algorithm as the most qualitative \textit{search-based} baseline. Our protocol involves comparisons across all the parameters of both the searching- and learning- based methods. 

Three different number of motifs will be computed $K \in \{3,10,30\}$ having two different lengths $L \in \{500, 1000\}$. Furthermore, the threshold $(T)$ of the experiments is chosen as a percentile in the distribution of distances between segments. To illustrate the setup, a length corresponding to the 1\%-th percentile, (denoted $\text{Pct}=1$ in Table~\ref{tab:bfmVsLm}) means that 1-\% of segments pairs have a pairwise Euclidean distance smaller than the threshold. In that way we can compare our method against the brute-force search across a range of thresholds computed by different percentiles $T \in \{ 0.001\%, 0.01\%,0.1\%, 1\% \}$ of the pairwise distances of segments. In that way we avoid hand-picking different thresholds values per dataset and select the threshold in a data-driven neutral manner. In order to ensure convergence, the learning rate was set to an initial value of $\eta=0.1$ and the number of iterations to $I=1000$. In addition, the optimization was restarted $R=200$ times. The segments were extracted from the series by sliding a window and normalizing the clipped segment, while the window is slid by half of the motif length. For every combination of the number of motifs $K$, length $L$ and threshold $T$ (computed from the percentile), three different values of frequency smoothness were searched $\alpha \in \{1,2,3\}$, keeping the one yielding the highest $\mathcal{F}$ value.

The brute-force search baseline was executed using the \textbf{same} $K,L,T (\text{Pct})$ combination parameters as the learning-based approach, and for both methods the final frequency $\mathcal{F}$ does \textbf{not} include trivial matches. In order to be entirely transparent to the research community we publicly shared our source code and the data used in this paper in an on-line repository\footnote{\url{http://fs.ismll.de/publicspace/LearnMotifs/}}. 

\subsection{Results}
\label{sec:empiricalResults}

\begin{table*}[htbp!]
  \centering
  \caption{Hard Frequencies ($\mathcal{F}(M)$): Learning Motifs (LM) vs. Brute Force Motifs (BFM)}
  \tabcolsep=0.06cm
    \begin{tabular}{|l||r|r!{\vrule width 1.1pt}r|r!{\vrule width 1.1pt}r|r!{\vrule width 1.1pt}r|r||r|r!{\vrule width 1.1pt}r|r!{\vrule width 1.1pt}r|r!{\vrule width 1.1pt}r|r|}
    \cline{2-17}
    \multicolumn{1}{c||}{} & \multicolumn{8}{c||}{\bf \textit{L=500}}                                     & \multicolumn{8}{c|}{\bf \textit{L=1000}} \\    \cline{2-17}
    \multicolumn{1}{c||}{} & \multicolumn{2}{c!{\vrule width 1.1pt}}{\bf \textit{Pct=0.001}} & \multicolumn{2}{c!{\vrule width 1.1pt}}{\bf \textit{Pct=0.01}} & \multicolumn{2}{c!{\vrule width 1.1pt}}{\bf \textit{Pct=0.1}} & \multicolumn{2}{c||}{\bf \textit{Pct=1}} & \multicolumn{2}{c!{\vrule width 1.1pt}}{\bf \textit{Pct=0.001}} & \multicolumn{2}{c!{\vrule width 1.1pt}}{\bf \textit{Pct=0.01}} & \multicolumn{2}{c!{\vrule width 1.1pt}}{\bf \textit{Pct=0.1}} & \multicolumn{2}{c|}{\bf \textit{Pct=1}} \\ \cline{2-17}
    \multicolumn{1}{c||}{} & \bf BFM   & \bf LM    & \bf BFM   & \bf LM    & \bf BFM   & \bf LM    & \bf BFM   & \bf LM    & \bf BFM   & \bf LM    & \bf BFM   & \bf LM    & \bf BFM   & \bf LM    & \bf BFM   & \bf LM \\ \cline{2-17} \hline \hline
\multicolumn{1}{|c||}{\bf \textit{Datasets}} & \multicolumn{8}{c||}{\bf \textit{Top-3 (K=3)}} & \multicolumn{8}{c|}{\bf \textit{Top-3 (K=3)}} \\ \hline
Insect B & 4 & \textbf{9} & 6 & \textbf{10} & 16 & \textbf{45} & 44 & \textbf{151} & 4 & \textbf{11} & 4 & \textbf{13} & 9 & \textbf{27} & 19 & \textbf{51} \\
TAO & 12 & \textbf{24} & 29 & \textbf{45} & 86 & \textbf{119} & 313 & \textbf{429} & 10 & \textbf{12} & 18 & \textbf{35} & 56 & \textbf{98} & 219 & \textbf{284} \\
RandomWalk & 25 & \textbf{43} & 74 & \textbf{125} & 239 & \textbf{321} & 697 & \textbf{855} & 9 & \textbf{23} & 27 & \textbf{64} & 114 & \textbf{165} & 327 & \textbf{458} \\
EEG LSF5 & 17 & \textbf{42} & 47 & \textbf{101} & 150 & \textbf{199} & 388 & \textbf{442} & 11 & \textbf{34} & 27 & \textbf{73} & 96 & \textbf{125} & 232 & \textbf{238} \\
Salinity & 39 & \textbf{48} & 151 & \textbf{184} & 497 & \textbf{590} & 1462 & \textbf{1718} & 18 & \textbf{32} & 72 & \textbf{94} & 269 & \textbf{330} & 683 & \textbf{876} \\
EOG & 153 & \textbf{190} & 504 & \textbf{669} & 1646 & \textbf{2168} & 4957 & \textbf{8042} & 67 & \textbf{102} & 196 & \textbf{340} & 676 & \textbf{1390} & 2171 & \textbf{5998} \\  \hline \hline
\multicolumn{1}{|c||}{\bf \textit{Datasets}} & \multicolumn{8}{c||}{\bf \textit{Top-10 (K=10)}} & \multicolumn{8}{c|}{\bf \textit{Top-10 (K=10)}} \\ \hline
Insect B & 11 & \textbf{18} & 14 & \textbf{23} & 35 & \textbf{78} & 81 & \textbf{189} & 11 & \textbf{29} & 11 & \textbf{28} & 17 & \textbf{54} & 46 & \textbf{97} \\
TAO & 30 & \textbf{48} & 62 & \textbf{95} & 192 & \textbf{314} & 780 & \textbf{1164} & 18 & \textbf{29} & 44 & \textbf{55} & 112 & \textbf{203} & 344 & \textbf{584} \\
RandomWalk & 40 & \textbf{79} & 132 & \textbf{206} & 313 & \textbf{579} & 1314 & \textbf{1502} & 23 & \textbf{48} & 52 & \textbf{118} & 223 & \textbf{310} & 603 & \textbf{768} \\
EEG LSF5 & 42 & \textbf{109} & 131 & \textbf{273} & 400 & \textbf{557} & 1118 & \textbf{1266} & 32 & \textbf{96} & 84 & \textbf{212} & 234 & \textbf{379} & 634 & \textbf{810} \\
Salinity & 100 & \textbf{105} & 291 & \textbf{358} & 1000 & \textbf{1149} & 2797 & \textbf{2995} & 47 & \textbf{59} & 136 & \textbf{198} & 456 & \textbf{597} & 1222 & \textbf{1564} \\
EOG & 263 & \textbf{283} & 973 & \textbf{1296} & 3128 & \textbf{4130} & 11181 & \textbf{13439} & 122 & \textbf{164} & 417 & \textbf{685} & 1552 & \textbf{2206} & 4321 & \textbf{5729} \\ \hline \hline
\multicolumn{1}{|c||}{\bf \textit{Datasets}} & \multicolumn{8}{c||}{\bf \textit{Top-30 (K=30)}} & \multicolumn{8}{c|}{\bf \textit{Top-30 (K=30)}} \\ \hline
Insect B & 31 & \textbf{40} & 36 & \textbf{47} & 68 & \textbf{107} & 200 & \textbf{221} & 32 & \textbf{49} & 32 & \textbf{49} & 42 & \textbf{72} & 89 & \textbf{110} \\
TAO & 65 & \textbf{95} & 133 & \textbf{209} & 432 & \textbf{698} & 1720 & \textbf{2193} & 38 & \textbf{55} & 65 & \textbf{93} & 202 & \textbf{336} & 577 & \textbf{932} \\
RandomWalk & 61 & \textbf{117} & 158 & \textbf{279} & 471 & \textbf{764} & 1778 & \textbf{2249} & 45 & \textbf{87} & 83 & \textbf{174} & 256 & \textbf{421} & 989 & \textbf{1151} \\
EEG LSF5 & 110 & \textbf{281} & 275 & \textbf{646} & 850 & \textbf{1442} & 2541 & \textbf{3505} & 72 & \textbf{205} & 153 & \textbf{428} & 417 & \textbf{879} & 1304 & \textbf{1914} \\
Salinity & 162 & \textbf{199} & 428 & \textbf{540} & 1260 & \textbf{1456} & 3270 & \textbf{3855} & 91 & \textbf{107} & 233 & \textbf{284} & 660 & \textbf{779} & 2038 & \textbf{2150}  \\
EOG & 427 & \textbf{557} & 1494 & \textbf{2028} & 5200 & \textbf{5681} & \textbf{17442} & 17075 & 247 & \textbf{338} & 787 & \textbf{1186} & 2306 & \textbf{2955} & 6227 & \textbf{7349} \\
\hline \hline
    \textbf{Wins} & 0 & \textbf{\underline{18}} & 0 & \textbf{\underline{18}} & 0 & \textbf{\underline{18}} & 1 & \textbf{\underline{17}} & 0 & \textbf{\underline{18}} & 0 & \textbf{\underline{18}} & 0 & \textbf{\underline{18}} & 0 & \textbf{\underline{18}} \\ \hline
    \end{tabular}%
  \label{tab:bfmVsLm}%
\end{table*}%

In the conducted experiments, for all the different thresholds $T$ (computed through the percentile), for all the different number of motifs $K$ and for different motif lengths $L$, the motifs learned through our method \textbf{almost always} had a higher frequency than the ones found through brute-force search. Table~\ref{tab:bfmVsLm} displays the empirical results comparing the frequency score of the optimal learned motif (denoted $LM$) against the motifs found through brute-force search (denoted $BFM$). 

The results of Table~\ref{tab:bfmVsLm} indicate that learning the motifs (LM) is better than searching (BFM) them in {$\bf 99.31\%$ of the experiments (143/144). The improvement arising from learning motifs (LM) in terms of motif frequencies is in average $ 67 \pm 56 \%$ better than the search-based approach (BFM). The famous Bland-Altman plot is used to assess the significance of the improvements. Figure~\ref{fig:blandAltman} (left plot) shows the dominating ratio of LM through least-squares fitting. Moreover, the right plot shows that the difference LM-BFM  and its standard deviations are above zero, thus we have a significant difference in terms of frequencies. 

\vspace{-0.3cm}
\begin{figure}[h]
\centering
\includegraphics[scale=0.43, trim=1.2cm 11cm 2.0cm 8cm]{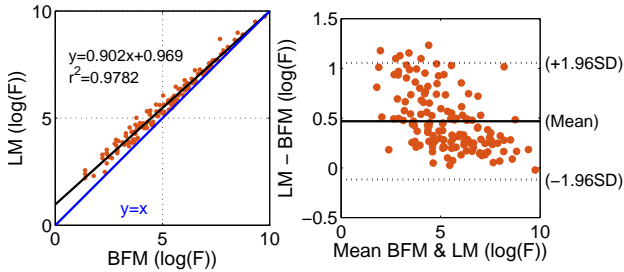}
\caption{Bland-Altman plot showing significance of LM vs BFM frequencies (log-scale for visual comprehension)}  
\label{fig:blandAltman}
\end{figure}

Even though the proposed method is significantly better in quality that the search-based alternatives, it is not the fastest method in the literature (which we never claimed). Yet, it is \textit{feasible} in terms of run-time, since learning the Top-30 motifs of Insect\_B (smallest dataset) took 4.7 minutes, while learning the Top-30 motifs of EOG (largest dataset) took 33.57 hours, in a cluster having Intel Xeon E5-2670v2 processors with speed 2.50GHz.

\section{Case Study: Audio Motifs}

In this case study we extract motifs from audio files. The case discussed in this thread is a poem by Edgar Allen Poe, titled "The  Bells" and famous for its onomatopoeic nature in terms of repeating the word "Bells". We extract a time-series representation of the audio file through the first channel of the Mel-frequency cepstral coefficients (MFCC). For the sake of illustration we took the first 300000 measurements of the original WAV file, corresponding to a 68 seconds audio reading of the poem. 

\begin{figure}[h!]
\centering
\includegraphics[scale=0.87, trim=1.8cm 8.6cm 0.0cm 6.8cm]{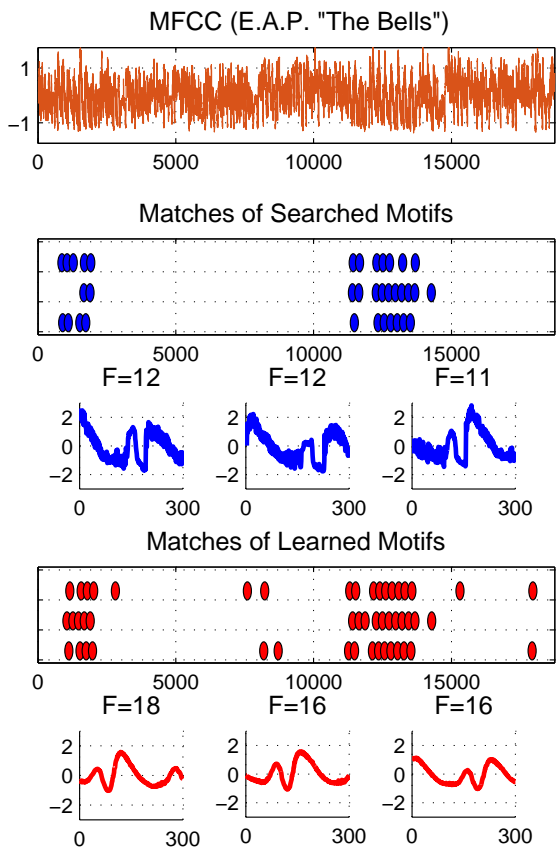}
\caption{Learning 3 audio motifs on a read version of the "The Bells" poem from Edgar Allan Poe. The method parameters are: $T=171.56\; (Pct=0.1\%), L=300, \alpha=3, \eta=0.3, I=1000, R=4$.}
\label{fig:bellsIllustration} 
\end{figure}

Figure~\ref{fig:bellsIllustration} illustrates shows the MFCC representation time-series together with the results of the brute force search algorithm in blue and our proposed method in red. We extracted three motifs $K=3$ of length $L=300$ for both methods. The distance threshold used in the experiment is the $0.1\%$-th percentile of pair-wise segment distances corresponding to a value of $T=171.56$.  For each method, we display the location of the motif matches over series segments with a filled oval mark. Under the plots of the matches we show the found motifs together with the corresponding frequencies. For the same distance threshold, the learned motifs have totally $50$ matches while the searched motifs have $35$ matches, for an improvement of $42\%$ in terms of frequency. Our method learns patterns that \textit{for exactly the same distance threshold} match more frequently than the brute-force motifs. 

An investigation of the motif sounds reveals that the top-K repetitive sounds are different pronunciations of the word bell. All the motifs are different from each other by 2T, so they are all legit motifs by definition. Let us analyze how optimality translates in concrete terms. For instance we can consider the segment between points 10000-15000 in the times series, which corresponds to the following poem text:

\begin{alltt}\normalfont
... Of the rapture that impels 
To the swinging and the ringing 
Of the \textit{bells}, \textit{bells}, \textit{bells} - 
Of the \textit{bells}, \textit{bells}, \textit{bells}, \textit{bells}, 
\textit{Bells}, \textit{bells}, \textit{bells} - 
To the rhyming and the chiming of the \textit{bells}! ...
\end{alltt}

Within the above segment, the brute-force motifs can find $7, 10, 7$ occurrences of the word bell within a threshold $T$. Our motifs can find $9, 11, 9$ matches within the same interval and for exactly the same distance threshold T. As the ground truth text above indicate, there are 11 "bells" pronunciations in total. In average, given the specified threshold $T$, the brute-force motifs find similar sounds that match to the word Bells in $72\%$ of the cases, the matches of our optimal motifs correspond to the word Bells in average on $88\%$ of the cases. This is a very important detection accuracy given that we used only the first channel of the MFCC representation, which is a low-resolution representation that encapsulates only the overall loudness of the sound.

\section{Conclusion}

This paper proposed a new perspective in learning time-series motifs. In contrast to current state of the art techniques which \textbf{searches} out motif candidates from series segments, our method \textbf{learns} them in a principled optimization. The motif frequency is approximated as a differentiable function and a gradient ascent method is proposed to find the motif values which maximize the objective function. In order to avoid local optima, a random restart strategy is combined with the gradient ascent learning of the motifs.

Learned optimal motifs have more segment matches than the motifs found through searching, for the same distance threshold. The optimal motifs represent latent patterns not necessarily present as sub-sequences in an explicit form, therefore can identify motifs which are in the center of the densest hyper-balls including segment points. Detailed experimental results demonstrate that learning optimal motifs \textbf{always} produces more qualitative motifs than searching them.

\section{Acknowledgment} 

This research has been co-funded by the Seventh Framework Programme of the European Commission, through the REDUCTION project (\#288254)\footnote{\url{www.reduction-project.eu}} and by Deutsche Forschungsgemeinschaft within the project HyLAP\footnote{\url{www.autonomous-learning.org}}. The authors also acknowledge Prof. Eamonn Keogh, University of California, Riverside, and Dr. Lucas Drumond, ISMLL, University of Hildesheim, whose suggestions helped improving the quality of the work.

\bibliographystyle{abbrv}
\bibliography{learnMotifs}

\end{document}